\documentclass[letterpaper]{article} % DO NOT CHANGE THIS
\usepackage{aaai25}  % DO NOT CHANGE THIS

\usepackage{times}  % DO NOT CHANGE THIS
\usepackage{helvet}  % DO NOT CHANGE THIS
\usepackage{courier}  % DO NOT CHANGE THIS
\usepackage[hyphens]{url}  % DO NOT CHANGE THIS
\usepackage{graphicx} % DO NOT CHANGE THIS
\usepackage{natbib}  % DO NOT CHANGE THIS AND DO NOT ADD ANY OPTIONS TO IT
\usepackage{caption} % DO NOT CHANGE THIS AND DO NOT ADD ANY OPTIONS TO IT
\usepackage{algorithmic}

\usepackage{multirow}
\usepackage{multicol}
\usepackage{booktabs}                 % 三线表-短细横线
\usepackage{amsfonts, amsmath, amssymb}
\usepackage[ruled, linesnumbered]{algorithm2e}
\usepackage{subfig}

\DeclareCaptionStyle{ruled}{labelfont=normalfont,labelsep=colon,strut=off} % DO NOT CHANGE THIS
\title{KOALA: Enhancing Speculative Decoding for LLM \\ via Multi-Layer Draft Heads with Adversarial Learning}

\author{
    Kaiqi Zhang,
    Jing Zhao,
    Rui Chen
}
\affiliations{
	Dalian University of Technology, China\\
    zhangkq@mail.dlut.edu.cn, zhaoj9988@dlut.edu.cn, 72117004@mail.dlut.edu.cn
}

\begin{document}

\maketitle

\begin{abstract}
	
Large Language Models (LLMs) exhibit high inference latency due to their autoregressive decoding nature. While the draft head in speculative decoding mitigates this issue, its full potential remains unexplored. In this paper, we introduce KOALA (K-layer Optimized Adversarial Learning Architecture), an orthogonal approach to the draft head. By transforming the conventional single-layer draft head into a multi-layer architecture and incorporating adversarial learning into the traditional supervised training, KOALA significantly improves the accuracy of the draft head in predicting subsequent tokens, thus more closely mirroring the functionality of LLMs. Although this improvement comes at the cost of slightly increased drafting overhead, KOALA substantially unlocks the draft head's potential, greatly enhancing speculative decoding. We conducted comprehensive evaluations of KOALA, including both autoregressive and non-autoregressive draft heads across various tasks, demonstrating a latency speedup ratio improvement of 0.24x-0.41x, which is 10.57\%-14.09\% faster than the original draft heads.

\end{abstract}

\section{Introduction}

\begin{figure}[t]
	\centering
	\includegraphics[width=0.48\textwidth]{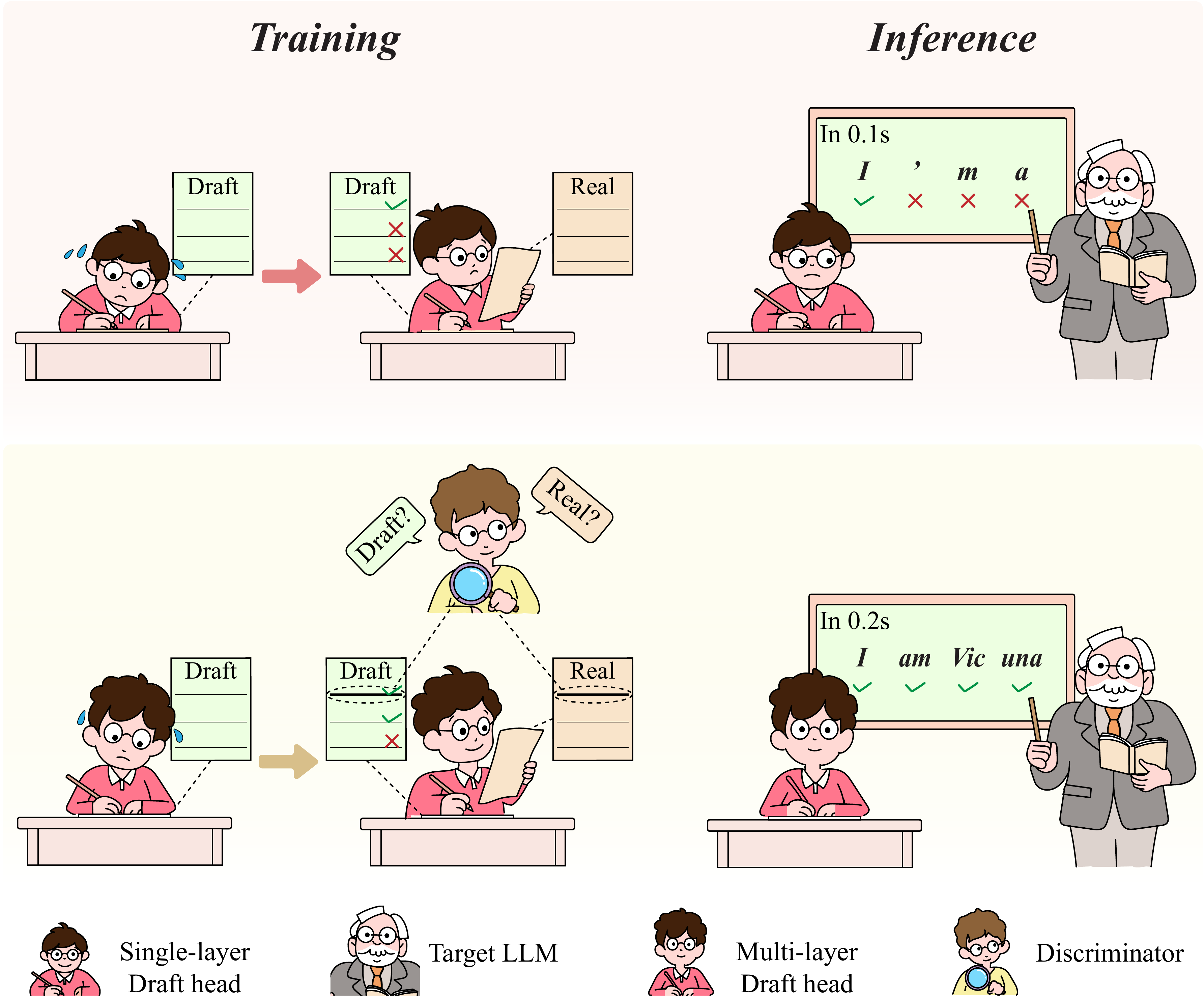}
	\caption{
		Comparison between the traditional draft head (upper panel) and the KOALA-optimized draft head (lower panel). KOALA expands the conventional single-layer structure to a multi-layer architecture and incorporates adversarial learning into traditional supervised training. While KOALA slightly increases drafting overhead, it substantially enhances speculative decoding efficiency by improving the draft head's accuracy in predicting subsequent tokens.
	}
	\label{overview of our method}
\end{figure}

Transformer-based \cite{vaswani2017attention} Large Language Models (LLMs), such as GPT-4 \cite{achiam2023gpt}, Llama 2 \cite{touvron2023llama}, and PaLM 2 \cite{anil2023palm}, demonstrate exceptional performance across various tasks. Due to their inherent autoregressive decoding nature, accelerating LLM inference has become a crucial research objective. Speculative decoding \cite{leviathan2023fast, chen2023accelerating}, utilizing a draft model, enhances the efficiency of target LLM inference through a \textit{draft-then-verify} paradigm. In each iteration of speculative decoding, the draft model initially predicts multiple subsequent tokens, which are then concurrently verified by the target LLM for acceptable continuations.

Speculative decoding hinges on finding a draft model that closely mirrors the target LLM’s functionality while achieving faster inference. Initial approaches employed \textit{independent drafting}, wherein a smaller, separate LM (e.g., T5-small) accelerates inference for a larger LM (e.g., T5-XXL). However, LMs from disparate series frequently exhibit incompatible implementation details, hindering interoperability. Moreover, the high costs of training a dedicated LM for speculative decoding constrain the practicality of independent drafting. Recent advancements introduce \textit{self-drafting} methods, which enhance LLM inference speed without relying on separate draft models. Numerous self-drafting techniques design lightweight draft models called \textit{draft heads}, leveraging the semantically rich hidden states of the target LLM. Draft heads can be classified into two categories based on their decoding approach: \textit{non-autoregressive} and \textit{autoregressive}. Medusa \cite{cai2024medusa} and EAGLE \cite{li2024eagle} are representative works in these respective domains.

Although draft heads achieve significant acceleration, several limitations persist:
1) Current draft heads employ a single-layer architecture, enabling rapid token prediction but resulting in a substantial performance gap compared to target LLMs due to parameter count disparity. This gap impedes effective collaboration between draft heads and target LLMs, limiting their potential.
2) Current draft head training methods rely on supervised learning, which only captures superficial input-output mappings. This approach inadequately enables draft heads to capture the underlying process for generating tokens consistent with the target LLM's output distribution, limiting their predictive accuracy.

To address these limitations and unlock the potential of draft heads, we introduce KOALA, an orthogonal technique for draft head optimization, as illustrated in Figure \ref{overview of our method}.
1) We propose a \textit{multi-layer} draft head structure to mitigate the performance gap with target LLMs caused by parameter disparities. In contrast to a single-layer design, this multi-layer architecture enables draft heads to more closely mirror target LLMs' functionality, enhancing overall collaboration.
2) We introduce a novel draft head training method that incorporates \textit{adversarial learning} into traditional supervised training. By leveraging the dynamic game mechanism between draft heads and discriminators, this approach encourages draft heads to better capture intricate token generation details in target LLMs, significantly improving prediction accuracy.
KOALA increases the number of tokens generated per draft-then-verify cycle, reducing the number of required algorithm iterations and enhancing speculative decoding efficiency. Notably, although the multi-layer structure slightly increases the draft overhead, it significantly accelerates the LLMs inference.

Our contributions can be summarized as follows:
\begin{itemize}
	\item We introduce KOALA, an orthogonal approach to improving draft head prediction accuracy that enhances speculative decoding efficiency. Specifically, KOALA includes two key innovations: expanding the conventional single-layer draft head into a multi-layer architecture and incorporating adversarial learning into traditional supervised training.
	
	\item We evaluated KOALA on the MT-bench using Medusa and EAGLE to represent non-autoregressive and autoregressive draft heads, respectively, with Vicuna models (7B, 13B, 33B) as target LLMs. Experimental results demonstrate that KOALA achieves a 0.24x-0.41x improvement in latency speedup ratio, which is 10.57\%-14.09\% faster than the original draft heads.
\end{itemize}

\section{Preliminaries}

\subsection{Autoregressive Decoding}

Autoregressive decoding is a fundamental technique employed by LLMs for sequence generation, wherein tokens are produced sequentially from left to right.

For an input sequence $x_1, x_2, \cdots, x_n$, the LLM $\mathcal{M}_q$ generates the subsequent token $x_{n+1}$ according to Equation \ref{pre: auto}.
\begin{equation} 
	\label{pre: auto} 
	x_{n+1} \sim q_{n+1} \leftarrow \mathcal{M}_q (x \, \vert \, x_{\leq n}) 
\end{equation}
Here, $q_{n+1}$ denotes the probability distribution of $x_{n+1}$ computed by $\mathcal{M}_q$, from which $x_{n+1}$ is sampled.

Subsequently, $x_{n+1}$ is appended to the input sequence, forming $x_1, x_2, \cdots, x_n, x_{n+1}$. This updated sequence is then fed into $\mathcal{M}_q$ to generate the next token $x_{n+2}$. This iterative process continues until $\mathcal{M}_q$ produces a complete sequence.

Given the autoregressive decoding nature of LLMs, accelerating their inference has emerged as a critical challenge.

\subsection{Speculative Decoding}

Speculative decoding employs a draft model to accelerate target LLM inference, while ensuring that the sampling results align with the target LLM. Speculative decoding adheres to a draft-then-verify paradigm. In each decoding iteration, the draft model initially predicts multiple future tokens efficiently, subsequently verified in parallel by the target LLM \cite{xia2024unlocking}.

For an input sequence $x_1, x_2, \cdots, x_n$, the draft model $\mathcal{M}_d$ efficiently predicts the subsequent $t$ tokens, as depicted in Equation \ref{pre: spe}.
\begin{equation}
	\label{pre: spe}
	\bar{x}_1, \bar{x}_2, \cdots, \bar{x}_{t} \sim d_1, d_2, \cdots, d_t \leftarrow \mathcal{M}_d (x \, \vert \, x_{\leq n})
\end{equation}
Here, $\mathcal{M}_d$ encompasses various draft methods, including both autoregressive and non-autoregressive decoding. The probability distributions $d_1, d_2, \cdots, d_t$ of the draft tokens govern the sequential sampling of $\bar{x}_1, \bar{x}_2, \cdots, \bar{x}_{t}$.

The target LLM $\mathcal{M}_q$ then verifies $\bar{x}_1, \bar{x}_2, \cdots, \bar{x}_{t}$ in parallel. $\mathcal{M}_q$ initially computes $t+1$ probability distributions simultaneously, as illustrated in Equation \ref{pre: verify}.
\begin{equation}
	\label{pre: verify}
	q_1, q_2, \cdots, q_{t}, q_{t+1} \leftarrow \mathcal{M}_q (x \, \vert \, x_{\leq n}, \bar{x}_{\leq t})
\end{equation}

Subsequently, each $\bar{x}_i$ undergoes an acceptance evaluation with a probability of $\min \left(1, {q_i(\bar{x}_i)}/{d_i(\bar{x}_i)} \right)$. The first rejected token, $\bar{x}_{f}$, is resampled using the adjusted distribution $\text{norm} (\max (0, q_f - d_f))$, while subsequent tokens $\bar{x}_{f+1}, \cdots, \bar{x}_{t}$ are discarded. If all tokens $\bar{x}_1, \cdots, \bar{x}_{t}$ are accepted, an additional token is sampled from $q_{t+1}$. The accepted tokens $\bar{x}_1, \cdots, \bar{x}_f$ are appended to the input sequence, creating the updated sequence $x_1, x_2, \cdots, x_{n}, \bar{x}_1, \cdots, \bar{x}_f$. This iterative process continues until the specified termination condition is met.

\subsection{Adversarial Learning}

Adversarial learning \cite{goodfellow2014generative} is a machine learning paradigm that primarily involves two components: a generator ($\mathcal{G}$) and a discriminator ($\mathcal{D}$). This learning framework enhances the realism of $\mathcal{G}$-generated data by enabling the two models to compete, co-evolve, and strive towards a Nash equilibrium during training. The objective of $\mathcal{G}$ is to produce realistic data, whereas $\mathcal{D}$ aims to differentiate between generated and authentic data.

In this framework, $\mathcal{G}$ generates data $\tilde{x}$ from input $z$, expressed as $\tilde{x} \leftarrow \mathcal{G}(z)$. $\mathcal{D}$ processes both authentic data $x$ and generated data $\tilde{x}$, outputting probabilities $\mathcal{D}(x)$ and $\mathcal{D}(\tilde{x})$ respectively, which indicate the likelihood of $\mathcal{D}$ classifying $x$ and $\tilde{x}$ as authentic.

The primary objective of adversarial learning is to train $\mathcal{G}$ to generate data so convincingly realistic that $\mathcal{D}$ cannot differentiate it from authentic data. This objective is realized through the optimization of the adversarial loss function, as depicted in Equation \ref{pre: adv_loss}.
\begin{equation}
	\label{pre: adv_loss}
	\min_{\mathcal{G}} \max_{\mathcal{D}} \mathbb{E}_{x \sim p_{\text{data}}(x)}[\log \mathcal{D}(x)] + \mathbb{E}_{z \sim p_z(z)}[\log (1 - \mathcal{D}(\mathcal{G}(z)))]
\end{equation}
Here, $\mathcal{D}$ strives to maximize the probability of correctly classifying authentic and generated data, whereas $\mathcal{G}$ attempts to minimize $\mathcal{D}$'s ability to differentiate between the two.

\section{KOALA}

KOALA optimizes the draft head in speculative decoding through its distinct structure and training process. To demonstrate KOALA, we employed Medusa and EAGLE as representatives of non-autoregressive and autoregressive draft heads, respectively.

\subsection{Multi-Layer Draft Head}

\begin{figure}[!t] 
	\centering
	\includegraphics[width=0.47\textwidth]{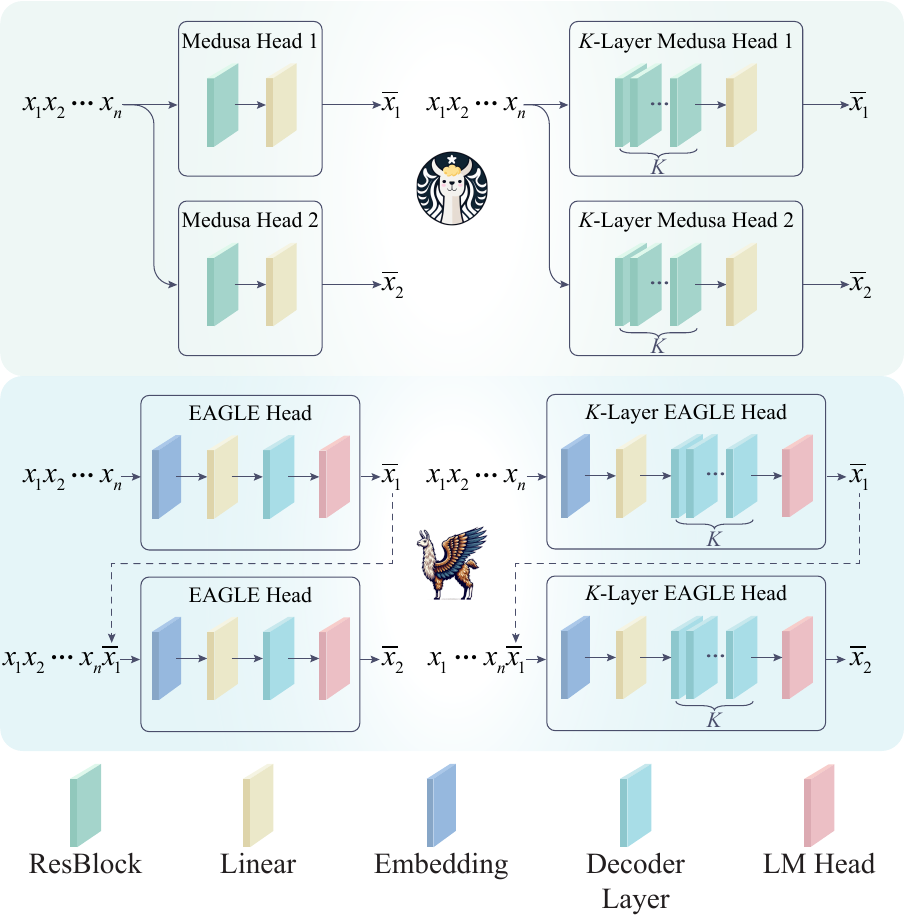}
	\caption{
		Comparison of single-layer and multi-layer draft head structures. For each Medusa Head, KOALA expands the single $\text{ResBlock}$ to $K$ layers. In the EAGLE Head, KOALA extends the single $\text{Decoder Layer}$ to $K$ layers. For simplicity, each draft head predicts only the next two tokens, $\bar{x}_1$ and $\bar{x}_2$, based on the input sequence $x_1, x_2, \cdots, x_n$.
	}
	\label{overview of multi layer}
\end{figure}

To reduce the performance gap between the draft head and the target LLM, KOALA transformed the single-layer draft head into a multi-layer structure, as illustrated in Figure \ref{overview of multi layer}.

The traditional Medusa Head comprises a Residual Block (ResBlock) followed by a Linear layer. The ResBlock predicts features of subsequent tokens, while the Linear layer maps these features to the vocabulary size. KOALA expanded this into a $K$-layer structure, represented as $(K \times \text{ResBlock} \rightarrow \text{Linear})$.

EAGLE Heads, in comparison, have a more complex structure. A conventional EAGLE Head consists of an Embedding, a Linear layer, a Decoder Layer, and an LM Head derived from the target LLM. The Embedding encodes historical tokens for autoregressive decoding, while the Linear layer integrates token and feature information before passing it to the Decoder Layer. The Decoder Layer then predicts features of subsequent tokens, which the LM Head maps to the vocabulary size. KOALA expanded this into a $K$-layer structure, represented as $(\text{Embedding} \rightarrow \text{Linear} \rightarrow K \times \text{Decoder Layer} \rightarrow \text{LM Head})$.

In summary, KOALA expands the single-layer draft head's prediction feature layer for subsequent tokens to $K$ layers, while maintaining the structure of other data processing and mapping layers. Notably, for LLMs with more transformer layers, indicating a larger performance gap with single-layer draft heads, a higher $K$ should be considered.

\subsection{Training with Adversarial Learning}

\begin{algorithm}[!t]
	\caption{Training Process for Draft Heads}
	\label{adversarial learning algorithm}
	\KwIn{Multi-Layer Draft head $\mathcal{M}_d$, Target LLM output logits $q$, Input sequence $x_1, x_2, \cdots, x_n$}
	\Repeat{$\mathcal{G}$ \text{and} $\mathcal{D}$ reach a Nash equilibrium}{
		\textbf{$\rhd$ Draft Head Step}\\
		\For{g-steps}{
			// \textit{$\mathcal{M}_d$ predicts logits for $t$ subsequent tokens}
			$d_{1}, d_2, \cdots, d_t \leftarrow \mathcal{M}_d (x \, \vert \, x_{\leq n})$\;
			// \textit{Draft Head Back Forward Pass}\\
			Compute $L_{\mathcal{G}} =  \underbrace{- \lambda \, \mathbb{E}_{\tilde{x} \sim p_{d}(d_{\leq t})} [ \log (\mathcal{D}(\tilde{x})) ]}_{\text{Adversarial Learning}}+ \underbrace{L_{\text{Distill}} (d_{\leq t}, q_{\leq t})}_{\text{Supervised Learning}} $\;
			Update draft head parameters\;
		}
		\textbf{$\rhd$ Discriminator Step}\\
		\For{d-steps}{
			// \textit{$\mathcal{M}_d$ predicts logits for $t$ subsequent tokens}
			$d_{1}, d_2, \cdots, d_t \leftarrow \mathcal{M}_d (x \, \vert \, x_{\leq n})$\;
			// \textit{Discriminator Back Forward Pass}\\
			Compute $L_{\mathcal{D}} = - \mathbb{E}_{\tilde{x} \sim p_{d}(d_{\leq t}) }[\log(1-\mathcal{D}(\tilde{x}))]$ $- \mathbb{E}_{\bar{x} \sim p_{q} (q_{\leq t})} \left[\log \mathcal{D}(\bar{x})\right] $\;
			Update discriminator parameters\;
		}
	}
\end{algorithm}

\begin{figure*}[!t] 
	\centering
	\includegraphics[width=\textwidth]{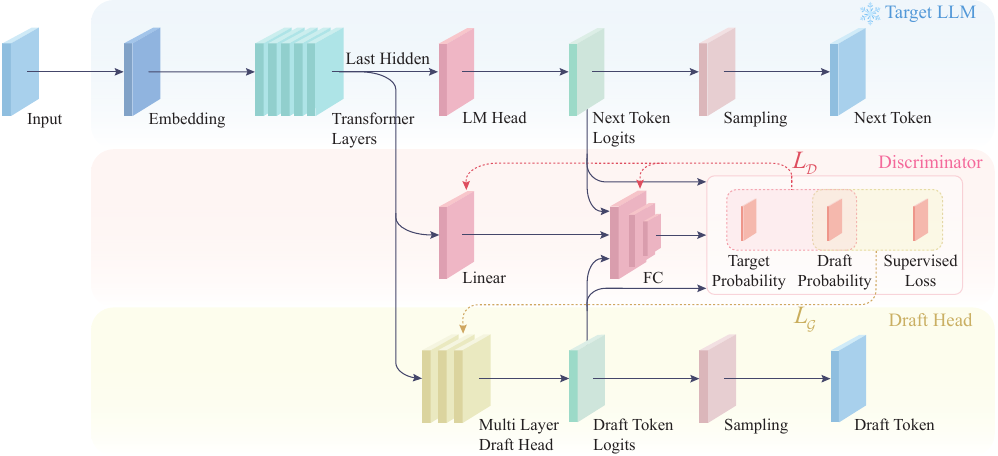}
	\caption{
		Training process for multi-layer draft heads which incorporates adversarial learning into supervised training. The target LLM, featuring a snowflake logo, and its parameters remain unupdated throughout the process. The discriminator and draft head are trained adversarially, co-evolving until they reach a Nash equilibrium, whereupon the training terminates.
	}
	\label{model}
\end{figure*}

To improve the draft head's token prediction accuracy, we integrate a discriminator into the training process, combining adversarial learning with supervised training.

In adversarial learning, the generator and discriminator co-evolve, necessitating comparable capabilities. To align capabilities and optimize training outcomes, we select discriminators with layer counts matching those of the draft head. Furthermore, the primary objective of draft head training is to mirror the target LLM's functionality. To further unlock the draft head's potential, we implement distillation rather than using a fixed dataset for supervised training, a method proven effective for training draft models in speculative decoding \cite{zhou2023distillspec}.

Figure \ref{model} illustrates the training process, comprising three main components: \textit{Target LLM}, \textit{Discriminator}, and \textit{Draft Head}. The Target LLM provides input and real data for draft head training without parameter updates. The Draft Head ($\mathcal{G}$) takes the semantically rich final hidden states of the Target LLM as input. After autoregressive or non-autoregressive decoding through the multi-layer draft heads, whose parameters are the only ones updated in $\mathcal{G}$ throughout the training process, draft token logits are obtained, and the predicted token is generated through sampling. The Discriminator ($\mathcal{D}$) consists of a linear layer and a fully connected layer (FC). First, the linear layer processes the last hidden states from the Target LLM, mapping them to the same dimension as the token logits. Subsequently, based on the mapped last hidden states, the next token logits from the Target LLM, and the draft token logits from the Draft Head, the FC computes the \textit{Target Probability} and \textit{Draft Probability}, which represent the likelihoods that the input token logits originate from the Target LLM and Draft Head, respectively. In addition, $\mathcal{D}$ also calculates the \textit{Supervised Loss} based on the next token logits and draft token logits, which serves as the supervised learning loss for distillation. Afterward, $\mathcal{D}$ updates its parameters based on the Target Probability and Draft Probability, while $\mathcal{G }$ updates its parameters using the Draft Probability and Supervised Loss. The loss functions $L_{\mathcal{G}}$ and $L_{\mathcal{D}}$ for $\mathcal{G}$ and $\mathcal {D}$ are presented in Equations \ref{LG} and \ref{LD}, respectively. 
\begin{equation}
	\label{LG}
	L_{\mathcal{G}} =  \underbrace{- \lambda \, \mathbb{E}_{\tilde{x} \sim p_{d}(d)} [ \log (\mathcal{D}(\tilde{x}))]}_{\text{Adversarial Learning}} + \underbrace{L_{\text{Distill}} (d, q)}_{\text{Supervised Learning}}
\end{equation}
\begin{equation}
	\label{LD}
	L_{\mathcal{D}} = - \mathbb{E}_{\tilde{x} \sim p_{d}(d) }[\log(1-\mathcal{D}(\tilde{x}))] - \mathbb{E}_{\bar{x} \sim p_{q} (q)} \left[ \log \mathcal{D}(\bar{x}) \right] 
\end{equation}
Here, $d$ and $q$ represent the tokens logits predicted by the draft head and generated by the target LLM, respectively. $ \lambda$ denotes the weight of the adversarial learning loss function in $L_{\mathcal{G}}$. $L_{\text{Distill}}(\cdot)$ represents the supervised learning loss function in distillation, such as cross-entropy loss.

Once $\mathcal{G}$ and $\mathcal{D}$ reach a Nash equilibrium, the training is deemed complete. Algorithm \ref{adversarial learning algorithm} summarizes the entire training process.

\section{Experiments}

\subsection{Experimental Setup}

To assess KOALA's efficiency, we utilize Medusa and EAGLE as representatives of non-autoregressive and autoregressive draft heads, respectively, with Vicuna models (7B, 13B, 33B) \cite{chiang2023vicuna} serving as target LLMs. Training utilizes the ShareGPT \cite{sharegpt} dataset with 68,000 dialogue iterations. Evaluations are performed on an A800 80G GPU using MT-Bench \cite{zheng2024judging}, a multi-turn conversation benchmark encompassing diverse tasks such as mathematical analysis, abstract extraction, and code generation. Unless otherwise specified, all experiments employ a greedy decoding strategy, accepting tokens only when they match the target LLM's greedy next-token generation.

Medusa and EAGLE layers are configured with $K$ = 1, 2, 3, while the discriminator's FC layers range from 1 to 3, with learning rates between [1e-5, 5e-4]. The adversarial learning loss function weight $\lambda$ in Equation \ref{LG} is set within the range [0.05, 0.5]. Both the draft head and discriminator are set to perform one iteration ($g$ = $d$ = 1). The evaluation is conducted with a batch size of 1. For fair comparison, the original Medusa and EAGLE are trained using knowledge distillation. All other parameters and training settings adhere to the original Medusa and EAGLE configurations. Additionally, since the discriminator introduced in KOALA has similar parameters to the draft head, the incorporation of adversarial learning in training approximately doubles the training cost compared to supervised training alone.

The following metrics are employed to evaluate KOALA:
\begin{itemize}
	\item Walltime speedup ratio: The speedup ratio achieved by the draft head compared to vanilla autoregressive decoding, serving as the primary performance metric.
	
	\item Average acceptance length $\ell$: The average number of tokens generated per forward pass by the target LLM equipped with the draft head. Higher $\ell$ values indicate improved draft head prediction accuracy.
	
	\item Acceptance rate $n$-$\alpha$: The draft head's accuracy in predicting the $n$th subsequent token. Following the original EAGLE settings, we use chain drafts without tree attention, evaluating the prediction accuracy for the first three tokens ($n$ = 1, 2, 3).
\end{itemize}

\subsection{Main Results}

\begin{figure}[t] 
	\centering
	\includegraphics[width=0.45\textwidth]{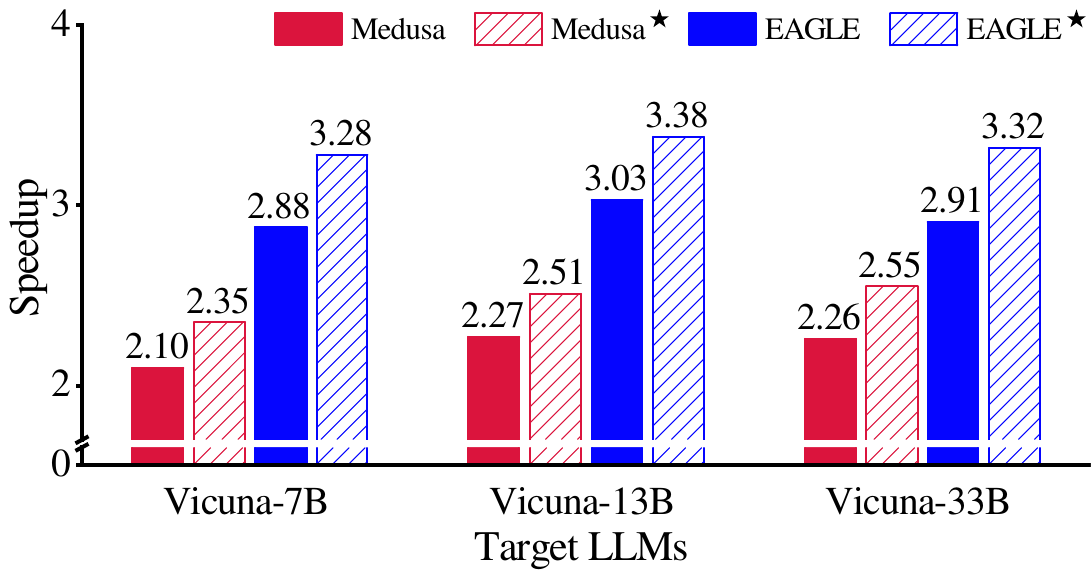}
	\caption{
		Speedup ratios of Medusa, EAGLE, and their KOALA-optimized versions achieving maximum speedup improvement, denoted by superscript $^{\bigstar}$. All configurations achieve maximum speedup at $K$ = 2, except Medusa on Vicuna-33B, which peaks at $K$ = 3.
	}
	\label{main result fig}
\end{figure}

Figure \ref{main result fig} and Table \ref{ex main table} demonstrate the effectiveness of KOALA.
We iterated through draft heads’ layers $K$ from 1 to 3 and reported the highest speedup ratio.
Compared with Medusa and EAGLE, representatives of non-autoregressive and autoregressive draft heads respectively, KOALA optimization improves the speedup ratio by 0.24x-0.29x and 0.35x-0.41x, which are 10.57\%-12.83\% and 11.55\%-14.09\% faster than their original draft heads.
%KOALA achieves a speedup ratio improvement of 0.24x-0.41x, outperforming the original draft head by 10.57\%-14.09\% for both Medusa and EAGLE. 
These results validate KOALA's efficacy for both non-autoregressive and autoregressive draft heads. The enhanced performance stems from the target LLM's increased acceptance rate of tokens predicted by the draft head. Specifically, the number of tokens generated per forward pass rises by 0.26-0.45, resulting in fewer iterations in the speculative decoding algorithm and consequently faster LLM inference.

\subsection{Ablation Study}

\subsubsection{Multi-Layer}

\begin{figure}[t] 
	\centering
	\includegraphics[width=0.45\textwidth]{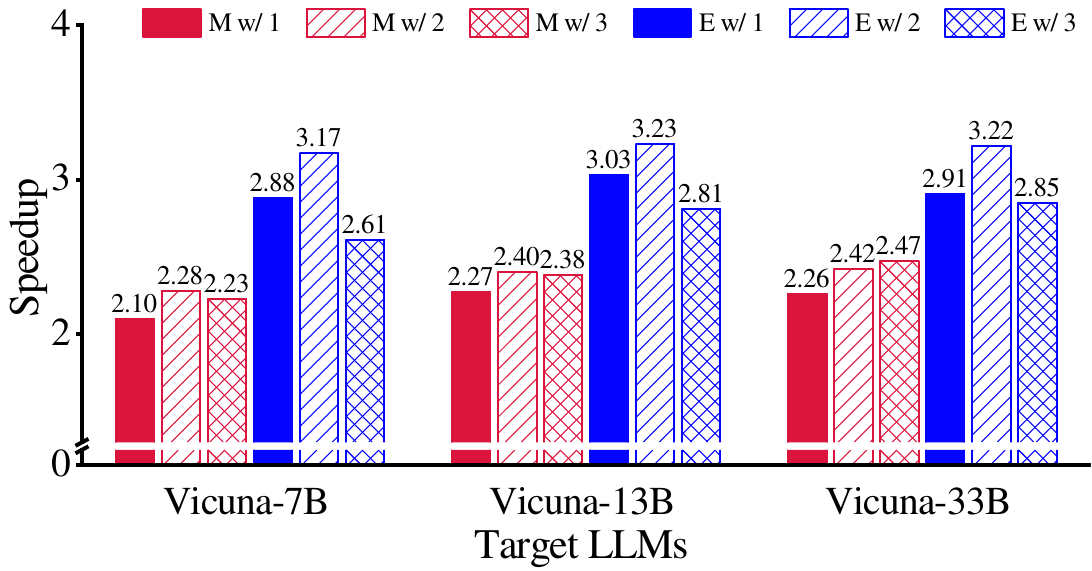}
	\caption{
		Speedup ratios of Medusa and EAGLE with varying layer structures. ``M w/ 1'' and ``E w/ 1'' represent the original single-layer Medusa and EAGLE, respectively.
	}
	\label{ml fig}
\end{figure}

KOALA transformers the traditional single-layer draft head into a multi-layer architecture. Figure \ref{ml fig} and Table \ref{ex main table} illustrate the performance comparison between multi-layer architecture ($K$ = 2, 3) and the original single-layer architecture ($K$ = 1), demonstrating the impact of using multi-layer approach.
Compared with the original single-layer Medusa and EAGLE,  the multi-layer architecture increases the average acceptance length by 0.18-0.45 and the speedup ratio by 0.11x-0.31x, indicating that the multi-layer architecture enables the draft head to better mirror the functionality of the target LLM. Notably, while the token acceptance rate and average acceptance length increase with $K$, the optimal speedup for most Medusa or EAGLE is achieved at $K$ = 2, with the exception of Medusa at $K$ = 3 on Vicuna 33B. This phenomenon is attributed to the increased number of draft head parameters in the multi-layer structure, which introduces additional drafting overhead. Consequently, it is crucial to balance the improved prediction accuracy against the increased drafting overhead by selecting an appropriate $K$. For Medusa and EAGLE, the multi-layer architecture achieves the most significant speedup improvements on the Vicuna 33B model, reaching 0.21x and 0.31x, respectively. This is attributed to the multi-layer architecture enhancing draft head performance by narrowing the parameter-induced performance gap between the draft head and the target LLM. Furthermore, in this experiment, the 33B model, containing the most transformer layers, exhibits the most pronounced performance disparity compared to the original single-layer draft head. Additionally, the speedup ratio of the draft head with $K$ = 3 improves as the target LLM size increases. Specifically, for Medusa, the speedup with $K$ = 3 shifts from near-optimal to optimal when moving from Vicuna 7B to Vicuna 33B. For EAGLE, although $K$ = 3 has not yet reached optimal performance, the gap is narrowing. We speculate that as the target LLM size further increases, EAGLE with $K$ = 3 or higher will yield optimal results. Consequently, higher $K$ values should be considered for larger target LLMs.


\begin{table*}[t]
	\centering
	\caption{
		Average acceptance lengths $\ell$ and acceptance rates $n$-$\alpha$ of Medusa, EAGLE, and their variants on Vicuna models.
		``V'' represents Vicuna.
		``M'' and ``E'' denote Medusa and EAGLE, respectively.
		``w/ AL'' indicates the draft head incorporating adversarial learning during training.
		``w/ 2'' and ``w/ 3'' signify draft heads using 2-layer and 3-layer architectures, respectively.
		The superscript $^{\bigstar}$ indicates the KOALA-optimized draft heads yielding the maximum speedup improvement in Figure \ref{main result fig}.
		We present the best results for Medusa and EAGLE series in \textbf{boldface}.
	}
	\label{ex main table}
	\begin{tabular}{cccccccccccc}
		\toprule
		& Model & Medusa & M w/ AL & M w/ 2 & M w/ 3 & Medusa$^{\bigstar}$ & EAGLE & E w/ AL & E w/ 2 & E w/ 3 & EAGLE$^{\bigstar}$ \\
		\midrule
		\multirow{3}{*}{$\ell$} 
		& V 7B 
		& 2.62 & 2.70 & 2.82 & 2.87 & \textbf{2.88} & 3.91 & 4.00 & 4.20 & \textbf{4.36} & 4.28 \\
		& V 13B 
		& 2.69 & 2.74 & 2.87 & 2.94 & \textbf{2.95} & 3.96 & 4.04 & 4.24 & \textbf{4.38} & 4.33 \\
		& V 33B 
		& 2.52 & 2.58 & 2.70 & 2.90 & \textbf{2.97} & 3.78 & 3.84 & 4.10 & \textbf{4.20} & 4.16 \\
		\midrule
		\multirow{3}{*}{1-$\alpha$} 
		& V 7B 
		& 0.57 & 0.58 & 0.59 & 0.60 & \textbf{0.60} & 0.79 & 0.79 & 0.81 & 0.82 & \textbf{0.82} \\
		& V 13B 
		& 0.58 & 0.59 & 0.61 & 0.62 & \textbf{0.63} & 0.79 & 0.79 & 0.81 & 0.83 & \textbf{0.83} \\
		& V 33B 
		& 0.55 & 0.57 & 0.59 & 0.61 & \textbf{0.64} & 0.77 & 0.78 & 0.80 & \textbf{0.81} & 0.81 \\
		\midrule
		\multirow{3}{*}{2-$\alpha$} 
		& V 7B 
		& 0.40 & 0.41 & 0.41 & 0.43 & \textbf{0.43} & 0.74 & 0.75 & 0.77 & \textbf{0.79} & 0.77 \\
		& V 13B 
		& 0.41 & 0.41 & 0.43 & \textbf{0.44} & 0.43 & 0.74 & 0.75 & 0.79 & \textbf{0.80} & 0.79 \\
		& V 33B 
		& 0.40 & 0.40 & 0.40 & \textbf{0.42} & 0.42 & 0.72 & 0.74 & 0.73 & \textbf{0.76} & 0.75 \\
		\midrule
		\multirow{3}{*}{3-$\alpha$} 
		& V 7B 
		& 0.31 & 0.31 & 0.32 & 0.32 & \textbf{0.34} & 0.69 & 0.71 & 0.74 & 0.75 & \textbf{0.75} \\
		& V 13B 
		& 0.31 & 0.33 & 0.33 & 0.34 & \textbf{0.34} & 0.70 & 0.71 & 0.75 & \textbf{0.76} & 0.75 \\
		& V 33B 
		& 0.30 & 0.30 & 0.31 & \textbf{0.32} & 0.32 & 0.68 & 0.67 & 0.69 & 0.71 & \textbf{0.71} \\
		\bottomrule
	\end{tabular}
\end{table*}

\subsubsection{Adversarial Learning}

\begin{figure}[t] 
	\centering
	\includegraphics[width=0.45\textwidth]{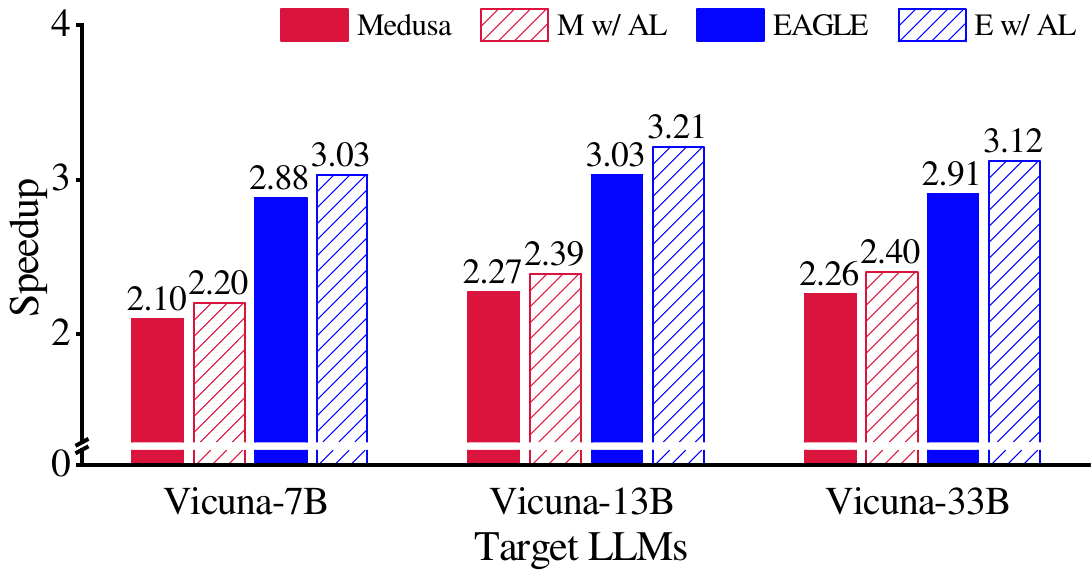}
	\caption{
		Speedup ratios of Medusa, EAGLE, and their variants incorporating adversarial learning during training.
	}
	\label{al fig}
\end{figure}

Another innovation of KOALA is the incorporation of adversarial learning into the conventional supervised training process for draft heads. Figures \ref{al fig} and Table \ref{ex main table} illustrate the comparative results, showcasing the impact of the adversarial learning approach. Compared to the original Medusa and EAGLE, the integration of adversarial learning increases the average acceptance length by 0.06-0.1 and improves the speedup ratio by 0.1x-0.19x. These enhancements indicate that adversarial learning effectively improves the prediction accuracy of draft heads, thereby enhancing speculative decoding. Notably, unlike the multi-layer structure, adversarial learning does not alter the original draft head architecture, thereby incurring no additional drafting overhead. Consequently, any enhancement in the draft head's prediction accuracy directly contributes to improved speedup performance. Interestingly, we observe that EAGLE demonstrates more substantial improvements compared to Medusa. This discrepancy may be attributed to the limited number of training epochs in Medusa's original configuration, potentially impeding the draft head and discriminator from reaching Nash equilibrium. Conversely, EAGLE's longer training period enables it to more fully exploit the potential of adversarial learning.

\subsection{Non-Greedy Decoding}

\begin{figure}[t] 
	\centering
	\includegraphics[width=0.45\textwidth]{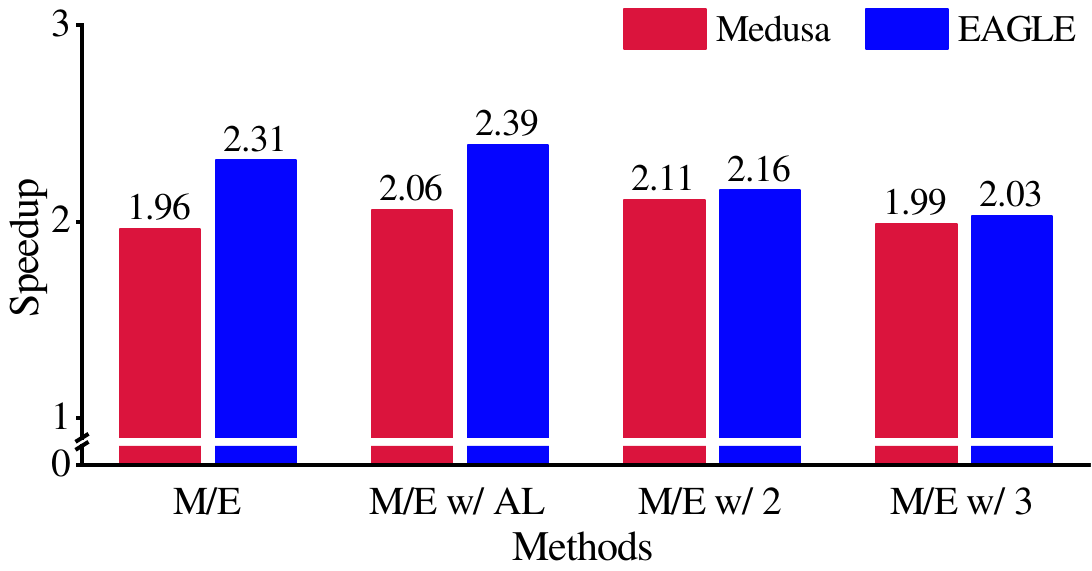}
	\caption{
		Speedup ratios of Medusa and EAGLE with various methods on Vicuna 7B under non-greedy settings.
		``M/E'' represents the original Medusa and EAGLE.
	}
	\label{t=1 fig}
\end{figure}

All evaluations thus far have been conducted under the greedy setting (temperature = 0). Figure \ref{t=1 fig} and Table \ref{table t=1} illustrate the evaluation results of KOALA under the non-greedy setting (temperature = 1). KOALA demonstrates diminished performance under non-greedy settings compared to greedy settings. For instance, for Vicuna 7B under greedy settings, the incorporation of adversarial learning achieves speedup ratio improvements ranging from 0.1x to 0.15x, while under non-greedy settings, they range from 0.08x to 0.1x. This observation also indicates that adversarial learning remains effective under the non-greedy setting, accelerating LLM inference by enhancing the draft head's prediction accuracy. Conversely, while the multi-layer structure improves Medusa's speedup, it adversely affects EAGLE's speedup ratio. 
\begin{table}[!t]
	\centering
	\caption{
		Average acceptance lengths $\ell$ of Medusa and EAGLE with various methods on Vicuna 7B under non-greedy settings.
	}
	\label{table t=1}
	\begin{tabular}{ccccc}
		\toprule
		& M/E & M/E w/ AL & M/E w/ 2 & M/E w/ 3 \\
		\midrule
		M &
		2.61 & 2.67 (+0.06) & 2.76 (+0.15) & 2.82 (+0.21) \\
		\midrule
		E &
		3.16 & 3.22 (+0.06) & 3.31 (+0.15) & 3.39 (+0.23)\\
		\bottomrule
	\end{tabular}
\end{table}
This discrepancy arises because, under the non-greedy setting, EAGLE's improvement in average acceptance length is minimal relative to its own baseline. However, the $K$-layer EAGLE introduces additional drafting overhead that increases with $K$, failing to balance the limited prediction accuracy improvement against the increased computational cost. Consequently, under the non-greedy setting, the implementation of the multi-layer structure should be context-dependent, considering the trade-offs between performance gains and drafting overhead.

\section{Related Work}

Recent studies aimed at enhancing the inference efficiency of LLMs have explored various techniques, including quantization \cite{frantar2022optq, dettmers2024qlora}, network pruning \cite{liu2023deja, frantar2023sparsegpt}, attention simplification \cite{chevalier2023adapting, zhang2024h2o}, and activation sharing \cite{shazeer2019fast, ainslie2023gqa}. These approaches aim to accelerate processing by reducing computational precision or minimizing the number of operations required. Furthermore, researchers have developed various strategies to optimize LLM inference architecture, such as non-autoregressive decoding \cite{stern2018blockwise, santilli2023accelerating}, early exiting \cite{xin2020deebert, zhou2020bert}, cascade inference \cite{wang2023tabi, chen2023frugalgpt}, and knowledge distillation \cite{taori2023stanford, chiang2023vicuna}. Although these techniques significantly accelerate LLM inference, they often involve trade-offs, as improvements in speed typically come at the expense of reduced generation quality.

Speculative decoding can achieve lossless acceleration through the draft-then-verify paradigm. Blockwise Decoding \cite{stern2018blockwise}, a pioneer of the draft-then-verify paradigm, introduces additional feedforward networks (FFNs) on top of the transformer decoder. This approach effectively accelerates greedy decoding by increasing generation parallelism. Subsequently, Speculative Sampling \cite{leviathan2023fast, chen2023accelerating} extends the concept from greedy decoding to non-greedy decoding methods. This technique demonstrates that the output distribution of speculative sampling remains consistent with that of the original sampling method.

Building upon these methods, researchers have explored various drafting approaches in speculative decoding, categorizing them into independent drafting and self-drafting techniques. SpecDec \cite{xia2023speculative} initially employs a non-autoregressive independent drafter, demonstrating significant acceleration effects. However, training the draft model from scratch incurs substantial computational costs. To reduce training costs, researchers have proposed using a smaller existing LM to accelerate a larger LM from the same series \cite{spector2023accelerating, sun2024spectr}. Nevertheless, coordinating LMs from different series remains challenging due to variations in their implementation details and architectural designs.

Self-drafting, which utilizes the target LLM for prediction, seamlessly integrates into existing systems without requiring an additional draft model. This approach effectively addresses the aforementioned challenges, demonstrating significant potential.  Recent research has extensively explored early exiting and layer skipping techniques within the target LLM for drafting purposes.  For instance, an additional early exit subprocess during decoding is introduced to predict the next token in advance \cite{yang2023predictive}. Likewise, several intermediate layers can be adaptively skipped during inference for efficient drafting \cite{zhang2023draft}.

Another promising research direction involves integrating lightweight non-autoregressive or autoregressive prediction heads after the target LLM's final hidden states, leveraging rich semantic information for next-token prediction. Medusa \cite{cai2024medusa} introduces multiple non-autoregressive draft heads after the final hidden states to generate candidate tokens in parallel, further exploiting the potential of FFN and advancing non-autoregressive methods. Amphista \cite{li2024amphista} enhances Medusa by introducing an automatic embedding block with a bidirectional self-attention module and a staged adaptation layer for feature transformation. Various complementary methods further exploit the potential of non-autoregressive draft heads. These include re-scoring algorithms based on local neural models and global n-gram models to optimize draft generation \cite{kim2024towards}, as well as multi-token prediction methods that simultaneously predict multiple future tokens during draft head training while maintaining consistent training time and memory overhead \cite{gloeckle2024better}. Hydra \cite{ankner2024hydra} leverages previously predicted token information to transform non-autoregressive draft heads into an autoregressive FFN. Clover \cite{xiao2024clover} enhances the prediction accuracy of regressive draft heads by incorporating sequential knowledge through regression connections, attention decoders, and enhancement modules. EAGLE \cite{li2024eagle} integrates token and feature information to transform the FFN into an autoregressive head, consisting of a fully connected layer and a decoder layer, thereby significantly improving the acceptance rate of draft tokens. Building upon EAGLE, EAGLE-2 \cite{li2024eagle2} dynamically adjusts the draft tree structure based on the confidence score of the draft model, further enhancing the inference efficiency of LLMs. Building upon existing draft head techniques, KOALA transforms the traditional single-layer draft head into a multi-layer structure and incorporates adversarial learning into conventional supervised training. This approach enables the draft head to more closely mirror the functionality of the target LLM, thereby enhancing speculative decoding.

\section{Conclusion}

In this paper, we introduce KOALA, an efficient orthogonal approach for draft head optimization that enhances speculative decoding for LLMs. KOALA transforms the traditional single-layer draft head into a multi-layer structure and incorporates adversarial learning into conventional supervised training. At the cost of a slight increase in drafting overhead, KOALA enables the draft head to more closely mirror the functionality of LLMs, thereby accelerating LLM inference. We conducted comprehensive evaluations of KOALA on Medusa and EAGLE, representing non-autoregressive and autoregressive draft heads, respectively, using Vicuna models (7B, 13B, 33B) as target LLMs and MT-bench dataset for assessment. KOALA achieves a 0.24x-0.41x improvement in latency speedup ratio, which is 10.57\%-14.09\% faster than the original draft heads.

\bibliography{aaai25}

\end{document}